\documentclass[10pt,twocolumn,letterpaper]{article}

\usepackage{3dv}
\usepackage{times}
\usepackage{epsfig}
\usepackage{graphicx}
\usepackage{amsmath}
\usepackage{amssymb}
\usepackage{float}
\usepackage{subfig}
\usepackage{pifont}
\usepackage[pagebackref=true,breaklinks=true,letterpaper=true,colorlinks,bookmarks=false]{hyperref}

\threedvfinalcopy % *** Uncomment this line for the final submission

\def\threedvPaperID{****} % *** Enter the 3DV Paper ID here

\ifthreedvfinal\fi
\begin{document}

\newcommand{\edit}[1]{{\color{blue}{#1}}}
\newcommand{\note}[1]{{\color{red}{#1}}}

%%%%%%%%% TITLE
\title{3DTextureTransformer: Geometry Aware Texture Generation for Arbitrary Mesh Topology}

\author{Dharma KC\\
The University of Arizona\\
Tucson, AZ, USA\\
{\tt\small kcdharma@arizona.edu}
\and
Clayton T. Morrison\\
The University of Arizona\\
Tucson, AZ, USA\\
{\tt\small claytonm@arizona.edu}
}

\maketitle
% \thispagestyle{empty}

%%%%%%%%% ABSTRACT
\begin{abstract}
   Learning to generate textures for a novel 3D mesh given a collection of 3D meshes and real-world 2D images is an important problem with applications in various domains such as 3D simulation, augmented and virtual reality, gaming, architecture, and design. Existing solutions either do not produce high-quality textures or deform the original high-resolution input mesh topology into a regular grid to make this generation easier but also lose the original mesh topology. In this paper, 
   we present a novel framework called the 3DTextureTransformer that enables us to generate high-quality textures without deforming the original, high-resolution input mesh. Our solution, a hybrid of geometric deep learning and StyleGAN-like architecture, is flexible enough to work on arbitrary mesh topologies and also easily extensible to texture generation for point cloud representations. Our solution employs a message-passing framework in 3D in conjunction with a StyleGAN-like architecture for 3D texture generation. The architecture achieves state-of-the-art performance 
   %\note{[update-results]} and establishes
   %\edit{Hi Professor, I am not clear about the update results part here}
   among a class of solutions that can learn from a collection of 3D geometry and real-world 2D images while working with
   any arbitrary mesh topology.
\end{abstract}

%%%%%%%%% BODY TEXT
\section{Introduction}
Texture generation for a given 3D mesh is an extremely important task with its applications in multiple domains like 3D simulation, augmented and virtual reality, gaming, design, and architecture. Although the field of 3D mesh generation from a text prompt and a single image is developing quite rapidly~\cite{wang2024prolificdreamer, poole2022dreamfusion, lin2023magic3d, wang2023score, metzer2023latent, chen2023fantasia3d, tsalicoglou2023textmesh, liu2023zero, qian2023magic123, kerbl20233d}, there has been less activity
in sampling new textures for these generated meshes. However, sampling new textures for these 3D meshes allows the user to make the scene realistic. It allows using differently textured meshes based on the context while also saving a lot of manual work required for manual texture generation. Recently, GGAN~\cite{dharma2022texture} and Texturify~\cite{siddiqui2022texturify} propose a framework based on Generative Adversarial Networks (GANs)~\cite{goodfellow2020generative} and Differentiable Rendering~\cite{kato2018neural, liu2019soft, ravi2020accelerating} for learning to generate textures on 3D meshes from a collection of 2D images. While GGAN works for arbitrary 3D mesh topology, it is not powerful enough for high-quality texture generation and Texturify deforms the original high-quality input mesh losing some important details and thus changes the topology of the input 3D mesh~\cite{yu2023texture}.
In this paper, we present a framework that builds upon GGAN and Texturify while removing their limitations, and has the following properties:
\begin{itemize}
\item The input mesh does not require any changes to the mesh topology or resolution.
\item It does not require ground truth textured 3D meshes, unlike~\cite{yu2023texture} and also does not require expensive 3D part segmentation unlike~\cite{gao2021tm, pavllo2021learning}.
\item The model is inspired by the Perceiver~\cite{jaegle2021perceiver} architecture.
\item It builds a Unet-like architecture based solely on the geometric information and thus is easily adaptable to newer datasets, unlike Texturify, which requires complicated data preprocessing.
\item It is easily extensible to point clouds and Gaussian splats~\cite{kerbl20233d}, while easily scaling to larger meshes and also works for both triangular and quad meshes.
\end{itemize}
As we demonstrate, the framework is a general, powerful and flexible generative model for 3D data. The code is made available open source.

%Our work builds upon GGAN and Texturify by removing their limitations and thus can produce high-quality textures on input mesh without any mesh deformation and changes to the mesh topology. Our solution does not require ground truth textured 3D meshes unlike~\cite{yu2023texture} and also does not require expensive 3D part segmentation unlike~\cite{gao2021tm, pavllo2021learning}. Our solution builds Unet-like architecture based solely on the geometric information and thus is easily adaptable to newer datasets, unlike Texturify which requires complicated data pre-processing. Our solution is also easily extensible to point clouds and Gaussian splats~\cite{kerbl20233d} while being easily scalable to larger meshes. Our solution works out of the box for both manifold and non-manifold meshes. Moreover, our solution works for both triangular and quad meshes. Our solution is inspired by Perceiver~\cite{jaegle2021perceiver} where we aim to create a general, powerful, and flexible generative model for 3D data. We will open-source our source code.
%-------------------------------------------------------------------------

\section{Related Work}
We aim to generate novel, high-quality textures for a previously unseen 3D mesh model. The following research is related to our model.
%In the following section, we outline some of the related research:

\subsection{2D image generation}
One line of research related to our problem is the generation of 2D images that are aware of 3D geometry, such as EG3D~\cite{chan2022efficient}, GRAF~\cite{schwarz2020graf}, GIRAFFE~\cite{niemeyer2021giraffe}, pi-GAN~\cite{chan2021pi}, and GRAM~\cite{deng2022gram}. These approaches are mainly focused on generating multiview consistent 2D images that respect the underlying 3D geometry of the image as opposed to texture generation for a given 3D mesh model.

\subsection{3D texture generation}
 Texture Fields~\cite{oechsle2019texture} generate a texture for a 3D geometry based on 2D image input by regressing a continuous 3D function parameterized by a neural network. Our work is slightly different than this setting because we want to sample new textures for a given 3D mesh model without being conditioned on the input 2D image during inference. Another closely related model 
 is LTG~\cite{yu2021learning}, but this approach
 requires an explicit UV mapping that distorts the regions near the seams. Text2Mesh~\cite{michel2022text2mesh} optimizes both the geometry and texture for a single 3D mesh instance based on a text prompt rather than generating novel textures for a given 3D mesh model. Finally, our work is closely related to GGAN~\cite{dharma2022texture} and Texturify~\cite{siddiqui2022texturify}; below, in Section 3, we explain how we extend these architectures while removing their limitations.

\subsection{GNNs}
Graph neural networks (GNNs) are powerful models for learning from graph-structured data. Our solution first converts the input 3D mesh into a geometric graph and learns node relationships using a message-passing framework where a node gets some information from its neighbors and updates its state.
Consider a graph $\mathcal{G} = (\mathcal{V}, \mathcal{E})$, where $\mathcal{V}$ is the set of nodes and $\mathcal{E}$ is the set of edges. The $k^{th}$ message passing iteration of a GNN is given by~\cite{hamilton2020graph}:
\begin{equation}
\begin{split}
        h_v^{(k+1)} &= \texttt{update}^{(k)}(h_v^{(k)}, \texttt{aggregate}^{(k)}(h_u^{(k)}),\\\
        &= \texttt{update}^{(k)}(h_v^{(k)}, \boldsymbol{m}_{\mathcal{N}(v)}^{(k)}) \quad \forall u \in \mathcal{N}(v))
\end{split}
\end{equation}
% Where 
Our solution uses a multi-headed self-attention among its neighbors only to aggregate the features and thus utilizes the sparsity of the adjacency matrix to make the solution scalable to larger graphs. 

\subsection{Graph Pooling and Unpooling}
We treat input mesh as a geometric graph and develop a Unet-like architecture. Graph pooling and unpooling operations allow us to develop powerful, efficient, and flexible Unet-like architecture. Graph pooling allows us to look at the input mesh at various hierarchical scales by iteratively coarsening the
graph into a new graph of smaller size. There are mainly two types of hierarchical graph pooling operations~\cite{liu2022graph}:
\begin{itemize}
    \item Node Clustering Pooling: This solution considers graph pooling as a node clustering problem and learns a cluster assignment matrix that maps nodes from a higher level to a lower level as:
    \begin{align}
    \begin{split}
        C &= f(X^l, A^l)\\
        X^{'} &= C^TX\\
        A^{'} &= C^TAC
    \end{split}
    \end{align}
    Where $C \in \mathbf{R^{n_l * n_{l+1}}}$ is a cluster assignment matrix and $f$ can be any function such as $softmax(GNN(X^l, A^l))$~\cite{ying2018hierarchical}. Although powerful, node clustering pooling operations are computationally expensive for larger graphs.
    \label{section:node_drop}\item Node Drop Pooling: Node drop pooling either uses a scoring function to select nodes or uses a sampling based strategy for node selection. Once these important nodes are selected, information from neighboring nodes can be used for pooling operations. The operations can be summarized as follows~\cite{liu2022graph}:
    \begin{align}
        \begin{split}
            S &= f_{score}(X^l, A^l)\\
            idx &= f_{select}(S)\\
            X^{'}, A^{'} &= f_{coarsen}(X^l, A^l, S, idx)
        \end{split}
    \end{align} Our geometric graph encoder uses graph pooling operations based on these node drop pooling operations as they are not computationally expensive thus making our solution scalable to larger meshes.
\end{itemize}
Specifically, we explore various graph pooling and unpooling operations such as sampling and max pooling~\cite{qi2017pointnet, qi2017pointnet++}, voxel-based pooling and unpooling~\cite{wu2022point}, Graph-Unets~\cite{gao2019graph},  and EdgePooling~\cite{diehl2019towards}. %Graclus~\cite{dhillon2007weighted},

\begin{figure*}
%\fbox{\rule{0pt}{2in} \rule{.9\linewidth}{0pt}}
 \centering
   \includegraphics[width=0.9\textwidth, height=4in]{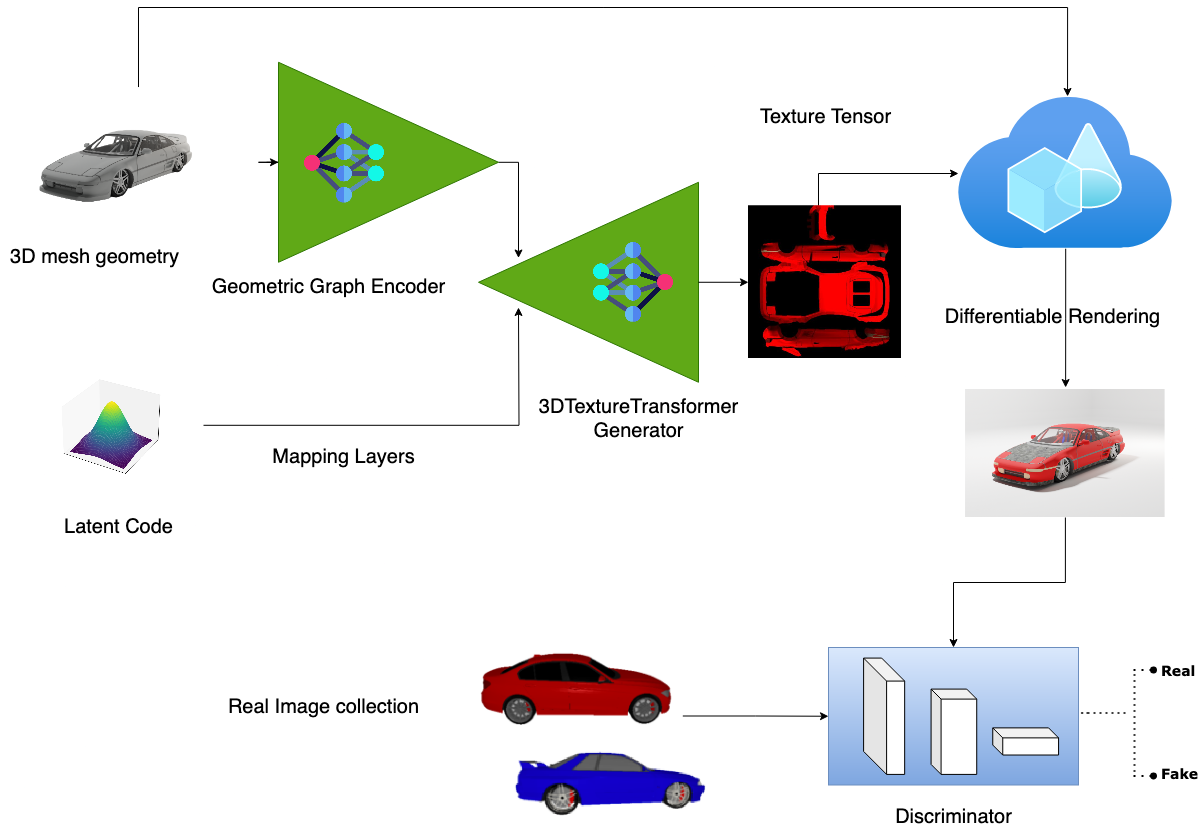}
   \caption{3DTextureTransformer Overall Pipeline}
\label{fig:3d_texture_transformer}
\end{figure*}

 \subsection{GANs and self-attention}
{\em Generative adversarial networks} (GANs)~\cite{goodfellow2020generative} are known for their ability to generate photorealistic images with very high resolution~\cite{karras2020analyzing}. The GAN framework consists of a generator $\mathcal{G}$ and discriminator $\mathcal{D}$. The generator tries to fool the discriminator by generating realistic samples while the discriminator tries to predict real samples from fake samples (generated by the generator). This constitutes a two-player minimax game with the following value function: 
\begin{equation}
    \begin{split}
        \mathcal{V}(\mathcal{G}, \mathcal{D}) &= \mathbb{E}_{\boldsymbol{x} \sim p_{data}(\boldsymbol{x})} [logD(\boldsymbol{x})] \\
        & + \mathbb{E}_{z\sim p_{z}(\boldsymbol{z})} [log(1 - \mathcal{D}(\mathcal{G}(\boldsymbol{z})))]
    \end{split}
\end{equation}

\noindent Here, $\boldsymbol{x}$ denotes a sample from a real distribution, $p_{data}$, and $\boldsymbol{z}$ denotes a ``noise'' vector from distribution $p_{z}$. StyleGAN~\cite{karras2020analyzing, karras2020training} is a state-of-the-art architecture for high-quality GAN based image synthesis with disentangled latent space. The StyleGAN architecture works perfectly with convolutional layers. As convolution requires a regular connectivity pattern, the architecture is limited to the case where we have a regular structure such as 2D images. Given that self-attention layers can handle arbitrary neighborhood structures and are able to handle long-range dependencies, there have been multiple attempts to combine self-attention layers with StyleGAN-like architectures. Recently, StyleFormer~\cite{park2022styleformer} and StyleSwin~\cite{zhang2022styleswin} use self-attention based generator architecture for 2D images. Our 3DTextureTransformer is a hybrid model that works with 3D data by utilizing self-attention layers with a message-passing framework in combination with StyleGAN-like architecture.

 \section{Methodology}
 The above figure~\ref{fig:3d_texture_transformer} shows the overall diagram of our system.

Our system learns to generate textures for a 3D mesh model from a collection of 3D mesh geometry and a collection of real-world 2D images. We use the same discriminator architecture as used by StyleGAN~\cite{karras2020training} and in the following section, we describe the geometric graph encoder and a generator called 3DTextureTransformer.

\begin{figure*}[h!]
    \begin{center}
%\fbox{\rule{0pt}{2in} \rule{.9\linewidth}{0pt}}
   \includegraphics[width=0.8\textwidth, height=5in]{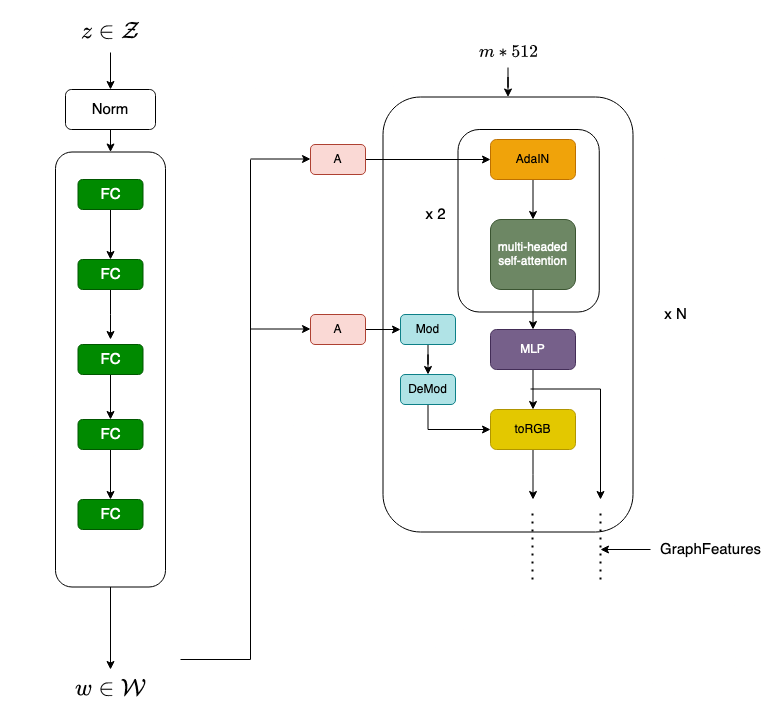}
   \caption{3DTextureTransformer Generator}
   \end{center}
\label{fig:generator}
\end{figure*}

\subsection{Geometric Graph Encoder}
Our geometric graph encoder utilizes node drop pooling~\ref{section:node_drop} operations to coarsen the geometric graph and its inverse operations for unpooling. We treat the centroid of a face in a mesh as a node in the geometric graph. We evaluate multiple strategies for pooling and unpooling the geometric graphs as follows:
\begin{itemize}
    \item FPSPooling: It uses farthest point sampling as a score function to select nodes in the given geometric graph~\cite{qi2017pointnet++}. Pooling is performed using max pool among its k nearest neighbors and unpooling is performed using interpolation among its k nearest neighbors~\cite{qi2017pointnet++} as follows:
    \begin{align}
    \begin{split}
        f(y) &= \frac{\sum_{i=1}^k w(x_i)f(x_i)}{\sum_{i=1}^k w(x_i)}\\
        w(x_i) &= \frac{1}{d(p(y), p(x_i))^2}
    \end{split}
    \end{align}
    Where d is the Euclidean distance, p(x) refers to its position, and f(x) refers to its features. 
    \item VoxelPooling: In voxel based pooling~\cite{simonovsky2017dynamic}, the 3D space is divided into regular grid structure called voxels. The faces of the triangular mesh that are within the same voxel are max pooled. For unpooling, we use the interpolation of all the points that lie within a given voxel.
    \item EdgePooling: EdgePooling contracts two pairs of nodes into a single node~\cite{diehl2019towards}. For our 3D mesh, this is equivalent to combining two faces into a single face. The contraction score is calculated for each edge as follows:
    \[r(e_{ij} = W(n_i||n_j) + b\]
    Where $n_i$ and $n_j$ are node features, and $W$ and $b$ are learnable parameters.
    \item GraphUnet: GraphUnet~\cite{gao2019graph} uses a learnable score function to select nodes in the graph. Once the nodes are selected, top-k nodes are selected and the information is aggregated from their neighbors. 
\end{itemize}
We evaluate the effectiveness of each approach for geometric graph encoding when combined with StyleGAN-like architecture for 3D texture generation.

\begin{figure}
\subfloat{\includegraphics[width=3in]{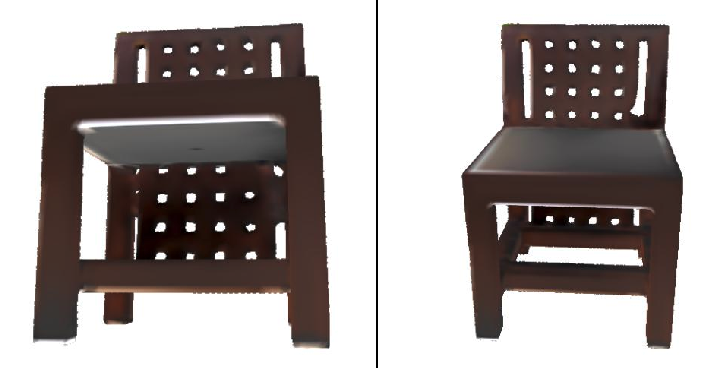}}\\
\subfloat{\includegraphics[width=3in]{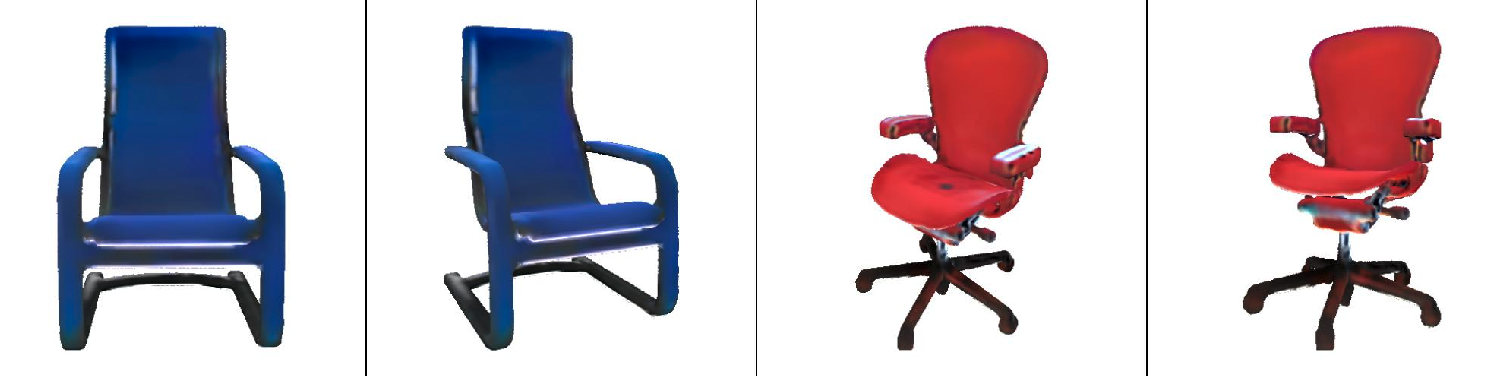}}\\
\subfloat{\includegraphics[width=3in]{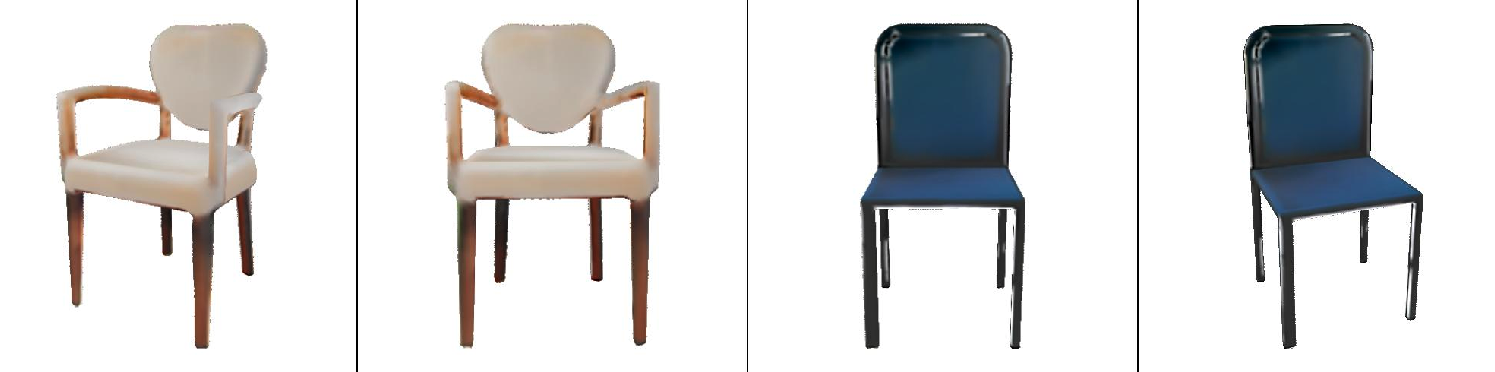}}\\
\subfloat{\includegraphics[width=3in]{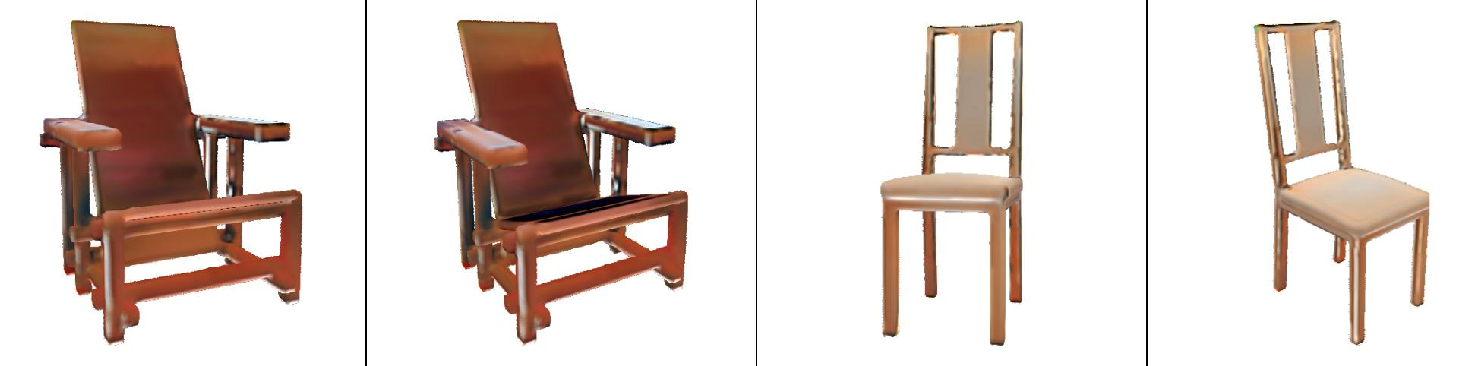}}\\
\subfloat{\includegraphics[width=3in]{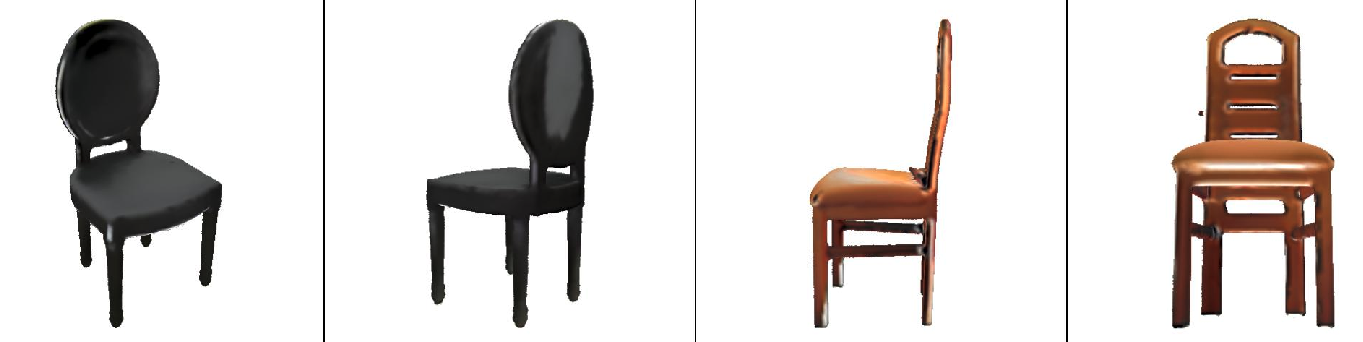}}\\
\caption{Textures generated by our solution on Chair Dataset}
\label{fig:results}
\end{figure}

\begin{table*}
\begin{center}
\begin{tabular}{|c|c|c|c|}
\hline
Method & Parameterization  & FID $\downarrow$ & KID * $10^-2$ $\downarrow$ \\%& Arbitrary Mesh topology \\
\hline\hline
Texture Fields~\cite{oechsle2019texture} & Global Implicit & 85.01 & 6.06  \\%& \cmark \\
\hline
SPSG~\cite{dai2021spsg} & Sparse 3D Grid & 65.36 & 5.13  \\%& \cmark \\
\hline
UV Baseline~\cite{siddiqui2022texturify} & UV & 38.98 & 2.46  \\%& \cmark \\
\hline
LTG~\cite{yu2021learning} & UV & 37.50 & 2.39  \\%& \cmark \\
\hline
EG3D~\cite{yu2021learning} & Tri-plane Implicit & 36.45 & 2.15 \\%& \cmark \\
%\hline
%Texturify~\cite{siddiqui2022texturify} & 4-RoSy Field & 26.17 & 1.54 & \xmark \\
\hline
\textbf{Ours} & \textbf{Sparse 3D implicit} & \textbf{33.87} & \textbf{1.93} \\%& \cmark \\
\hline
\end{tabular}
\end{center}
\caption{Results}
\label{table:results}
\end{table*}

\subsection{3DTextureTransformer}
 The above diagram \ref{fig:generator} shows our generator architecture. We use mapping layers to map latent code ($z \in \mathcal{Z}$) into smooth latent space ($w \in \mathcal{W}$). $A$ refers to the affine transformation of these smooth latent codes. Our network applies Adaptive instance normalization (AdaIN) operation on the features as follows~\cite{karras2019style}:
 \[AdaIN(x_i, y) = y_{s,i} * \frac{(x_i - \mathbb{E}{(x_i)})}{\sigma(x_i)} + y_{b,i}\]
 Where $s_i$ and $b_i$ are scaling and bias derived using affine transformation $A$ of the smooth latent space $w \in \mathcal{W}$ and are applied to $i^{th}$ input channel($x_i$). For toRGB layer, we use the weight modulation and demodulation operation as follows~\cite{karras2020analyzing}:
 \begin{align}
     w_{ij}^{'} &= s_i * w_{ij}
 \end{align}
 where $w$ and $w^{'}$ are original and modulated weights and $s_i$ is the scaling factor for $i^{th}$ input feature channel and $j$ enumerates the output feature maps. The demodulation is done using:
 \begin{align}
     w_{ij}^{''} &= \frac{w_{ij}^{'}}{\sqrt{\sum_i{w_{ij}^{'}}^2} + \epsilon}
 \end{align}
 The multi-headed local self-attention is calculated using the following sparse multi-headed operation~\cite{dwivedi2020generalization} making the solution scalable to larger graphs:
\[ SparseAttn(Q, K, V) = softmax(\frac{QK^T*A}{\sqrt{d}})V\]
Where Q, K, and V are Query, Key, and Value obtained from the linear transformation of input signal~\cite{vaswani2017attention} and $*$ denotes a Hadamard product with the sparse adjacency matrix $A$.

 The graph features encoded from the geometric graph encoder are concatenated with the features learned using the generator block. The advantage of such an architecture with AdaIn operations is that it allows experimentation with novel and powerful geometric deep-learning solutions in the future.

 \section{Experiments}
 We experiment with our solution in the Chairs and Cars dataset~\cite{siddiqui2022texturify}. This dataset uses a collection of 3D geometry from ShapeNet~\cite{chang2015shapenet} and a collection of 2D real-world chair images~\cite{park2018photoshape} and 2D real-world car images~\cite{yang2015large}. We evaluate our solution based on Frechet Inception Distance (FID)~\cite{heusel2017gans} and Kernel Inception Distance (KID)~\cite{binkowski2018demystifying} and follow the same evaluation pipeline of Texturify~\cite{siddiqui2022texturify}. We use PyTorch~\cite{paszke2019pytorch}, PyTorch Geometric~\cite{fey2019fast}, and DGL~\cite{wang2019deep} for all our experiments and Nvdiffrast~\cite{laine2020modular, siddiqui2022texturify} for differentiable rendering. We use a learning rate of 0.0002 for the generator, geometric graph encoder, and discriminator. Once the generator generates and applies the texture to a given mesh, it is rendered from 4 random camera viewpoints. We use positions, normals, gaussian curvature, and mean curvature as the input features for each face of the input 3D mesh.

\section{Results}
The above figure ~\ref{fig:results} shows some of the results from our system. The table~\ref{table:results} shows FID and KID values on the Chair Dataset and comparison with methods that can work with arbitrary mesh topology. Texturify~\cite{siddiqui2022texturify} does not work with arbitrary mesh topology and EG3D produces 2D images instead of texture that can be applied to 3D mesh.

\section{conclusion}
In this work, we have developed a hybrid architecture that can generate texture in 3D with a combination of geometric deep learning and StyleGAN-like architecture. Our solution uses self-attention layers in the 3D geometric space for message passing and thus makes our solution work on arbitrary mesh topology. Our solution uses a geometric graph encoder that is flexible enough to work on the 3D mesh, point clouds, and Gaussian splats. Currently, we observe that the textures generated have low frequency. In the future, we will explore a direction to add high-frequency details to these generated textures. And we will also explore the use of our solution for texture generation for point clouds and Gaussian splats in the future.

{\small
\bibliographystyle{ieee}
\bibliography{egbib}
}

\end{document}